\documentclass{article}

\usepackage[preprint]{neurips_2026}

\usepackage[utf8]{inputenc}
\usepackage[T1]{fontenc}
\usepackage{hyperref}
\usepackage{url}
\usepackage{booktabs}
\usepackage{amsfonts}
\usepackage{amsmath}
\usepackage{amssymb}
\usepackage{nicefrac}
\usepackage{microtype}
\usepackage{xcolor}
\usepackage{multirow}
\usepackage{tikz}
\usetikzlibrary{positioning, arrows.meta, fit, calc, backgrounds}

\title{ClaimDiff-RL: Fine-Grained Caption Reinforcement Learning through Visual Claim Comparison}

\author{
{\bf Tianle Li}$^{1}$ \quad
{\bf Xuyang Shen}$^{2}$ \quad
{\bf Yan Ma}$^{2}$ \quad
{\bf Rongxin Guo}$^{2}$ \quad
{\bf Shaoxiang Chen}$^{2}$ \quad
{\bf Jiacheng Chen}$^{1}$ \\
{\bf Haochen Wang}$^{2}$ \quad
{\bf Hongyang Tang}$^{2}$ \quad
{\bf Yucong Zhou}$^{2}$ \quad
{\bf Yu Cheng}$^{1}$ \\
\\
$^{1}$The Chinese University of Hong Kong \\
$^{2}$MiniMax \\
\texttt{tianleli@link.cuhk.edu.hk} \quad
\texttt{shenxuyang@minimaxi.com} \\
\texttt{chengyu@cse.cuhk.edu.hk} \\
\url{https://github.com/ltl3A87/ClaimDiff-RL}
}

\begin{document}

\maketitle

\begin{abstract}
Long-form image captioning exposes a reward granularity problem in RL: captions are judged as whole sequences, while the important errors occur at the level of individual visual claims. A good dense caption should be both faithful and informative, avoiding hallucination without omitting salient details. Yet pairwise preferences, reference-based metrics, and holistic scalar rewards compress these local errors into a single sequence-level signal, obscuring the tradeoff between factuality and coverage. We introduce \textsc{ClaimDiff-RL}, a framework that uses reference-conditioned atomic claim differences as the reward unit for caption RL. Given an image, an actor caption, and a reference caption, a multimodal judge enumerates visually grounded differences, verifies each difference against the image, assigns open-vocabulary error types and severity levels, and produces per-difference statistics for reward composition. This makes hallucinated claims and omitted salient facts separately measurable and tunable. Experiments show that holistic scalar rewards can reduce hallucination by increasing missing facts, while \textsc{ClaimDiff-RL} exposes this faithfulness and coverage tradeoff and enables more balanced operating points. On a 160-image human-labeled diagnostic benchmark, public captioning benchmarks, and VQA benchmarks, \textsc{ClaimDiff-RL} improves the hallucination--missing-fact balance, preserves general capability, and even surpasses Gemini-3-Pro-Preview on several fine-grained Capability dimensions such as object counting, spatial relations, and scene recognition. These results suggest that typed, verifiable claim differences are an effective reward unit for fine-grained and diagnosable caption RL.
\end{abstract}

\section{Introduction}
\label{sec:intro}

\begin{figure}[t]
\vspace{-2em}
    \centering
    \includegraphics[width=\textwidth]{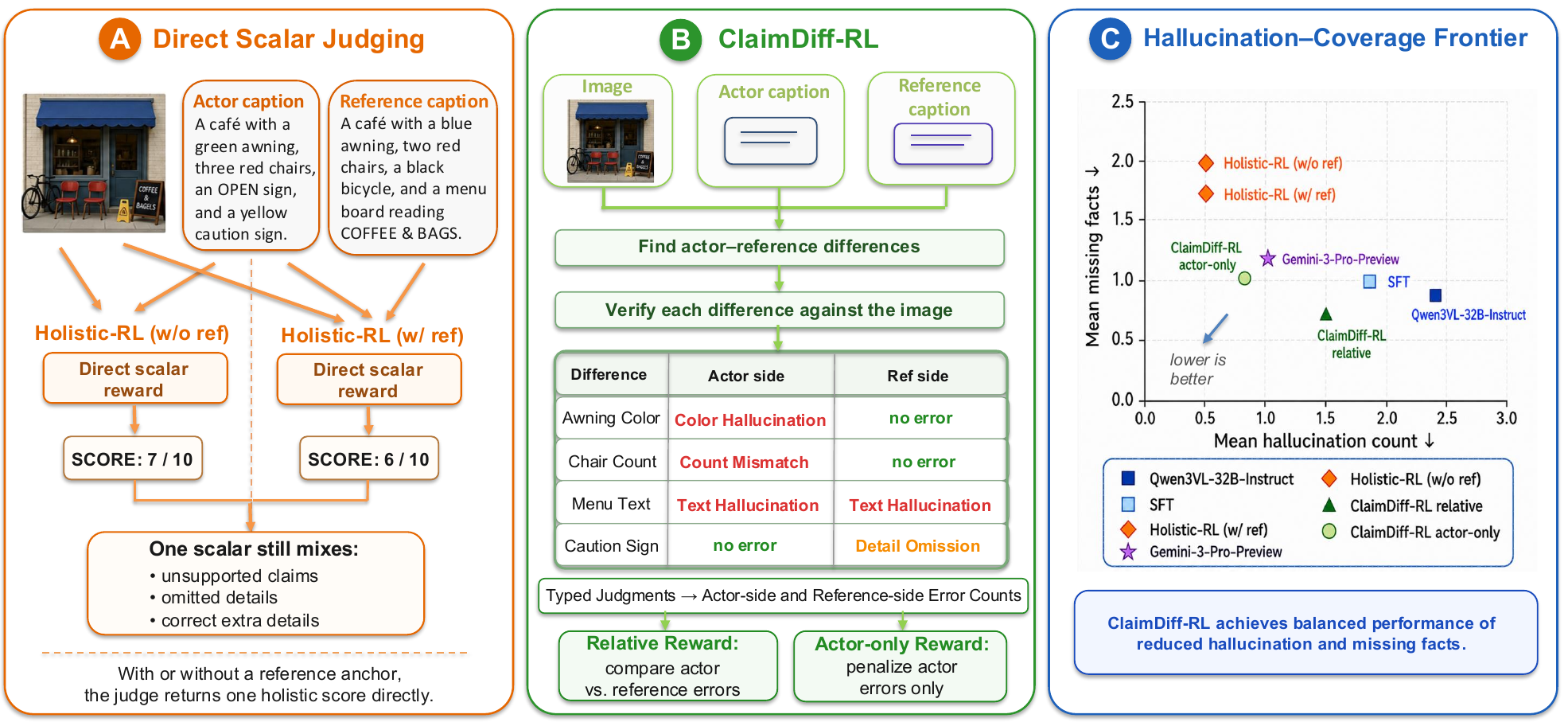}
    \caption{
    Overview of \textsc{ClaimDiff-RL}. Unlike direct scalar judging, \textsc{ClaimDiff-RL} verifies actor--reference visual differences against the image and composes typed side-specific errors into scalar rewards, making the hallucination--coverage tradeoff explicit.
    }
    \vspace{-0.5em}
    \label{fig:teaser}
\end{figure}

Long-form image captioning exposes a reward granularity problem in RL for open-ended generation. Unlike tasks where correctness can be summarized by a single answer, a dense caption is composed of many local visual claims about objects, attributes, counts, spatial relations, OCR text, identities, and fine-grained scene details. Earlier captioning objectives and metrics, such as CIDEr~\citep{vedantam2015cider}, SPICE~\citep{anderson2016spice}, and self-critical sequence training~\citep{rennie2017selfcritical}, made important progress by optimizing caption models toward reference-based evaluation signals. However, long-form captioning requires a more delicate objective than reference similarity alone. A caption can avoid hallucination by becoming overly conservative, or it can improve coverage by adding details that introduce unsupported claims. This tension is closely related to the hallucination problem studied in image captioning and LVLM evaluation~\citep{rohrbach2018object,li2023evaluating}. A good dense caption should therefore be both faithful and informative: it should avoid unsupported visual claims while still covering salient image content~\citep{wang2025vicritverifiablereinforcementlearning,zhong2025focusinternalmllmrepresentations}.

Most existing reward designs still score captions at the sequence level. Pairwise preference and RLHF-style methods compare complete outputs or learn holistic reward models~\citep{ouyang2022training,rafailov2023direct}; LLM-based caption evaluators such as CLAIR~\citep{chan2023clair} and MLLM-as-judge methods such as VIEScore and Prometheus-Vision~\citep{ku2024viescore,lee2024prometheusvision} show that strong foundation models can provide useful scalar judgments and explanations. Yet direct scalar judging remains opaque as a reward signal: a higher score does not reveal whether the caption became more visually grounded, less detailed, or simply safer. This issue remains even when a reference caption is provided. As illustrated in Figure~\ref{fig:teaser}, both \textsc{Holistic-RL} with a reference and \textsc{Holistic-RL} without a reference perform direct scalar judging; the only difference is whether the judge sees a comparison anchor. In both cases, hallucinations, missing facts, and correct extra details are compressed into one reward. Our experiments show that this compression can encourage conservative under-captioning, where hallucination is reduced by omitting more salient details.

Recent work has begun to move beyond monolithic caption scores. CapRL~\citep{xing2025caprl} defines caption quality through downstream utility, using whether a vision-free LLM can answer questions from the caption as a verifiable reward. SC-Captioner~\citep{zhang2025sccaptioner} decomposes predicted and reference captions into object, attribute, and relation sets using scene-graph parsing, and rewards self-correction by comparing the added and removed elements. These approaches suggest that caption rewards benefit from more structured supervision. However, utility-based rewards can still hide which visual claims caused success or failure, and fixed scene-graph schemas may miss open-ended visual dimensions such as OCR, style, identity, lighting, repetition, ambiguity, and fine-grained layout. The missing ingredient is not merely a stronger judge, but a better judging interface: one that turns global caption scoring into local, image-grounded verification before composing the scalar reward.

 We introduce \textsc{ClaimDiff-RL}, a caption RL framework that keeps the final reward compatible with standard scalar-reward optimization, but changes the reward unit from holistic caption scores to image-verified claim differences. Given an image, an actor caption, and a reference caption, a multimodal judge identifies actor--reference differences, verifies each difference against the image, assigns side-specific typed errors, and composes the resulting statistics into scalar rewards. The reference caption is used only as a comparison anchor, not as exhaustive ground truth.

    
    

Our contributions are threefold:
\begin{itemize}
    \item We propose \textbf{claim-difference judging} as a fine-grained reward interface for long-form caption RL. The judge identifies actor--reference visual differences, verifies them against the image, and assigns side-specific typed errors.

    \item We design \textbf{relative} and \textbf{actor-only} reward compositions from the same typed error statistics. These rewards expose different operating points on the faithfulness--coverage frontier.

    \item We show that holistic rewards often reduce hallucination by increasing omissions, while \textsc{ClaimDiff-RL} provides more controllable tradeoffs and preserves or improves captioning and VQA capability.
\end{itemize}

\section{Related Work}
\label{sec:related}

\paragraph{Automatic metrics for image captioning}
Image captioning has traditionally been evaluated with reference-based metrics such as BLEU~\citep{papineni2002bleu}, METEOR~\citep{banerjee2005meteor}, CIDEr~\citep{vedantam2015cider}, and SPICE~\citep{anderson2016spice}. These metrics provide scalable evaluation signals and have also been used as optimization targets, but they are poorly matched to long-form dense captioning, where many valid captions can differ in wording, order, length, and level of detail. Embedding-based or model-based metrics such as CLIPScore~\citep{hessel2021clipscore} and CAPTURE~\citep{dong2024benchmarking} move beyond surface overlap, and LLM or VLM-as-judge evaluators such as CLAIR~\citep{chan2023clair}, VIEScore~\citep{ku2024viescore}, and Prometheus-Vision~\citep{lee2024prometheusvision} provide stronger semantic judgments. However, these methods still often aggregate caption quality into a holistic score, making it difficult to tell whether a score reflects fewer hallucinations, better coverage, or simply safer and shorter descriptions.

\paragraph{Fine-grained diagnosis of caption quality}
Recent evaluation work increasingly treats caption quality as a collection of local visual claims rather than a single sentence-level property. Hallucination-focused metrics and benchmarks such as CHAIR~\citep{rohrbach2018object}, POPE~\citep{li2023evaluating}, HallusionBench~\citep{guan2024hallusionbench}, and MMHal-Bench~\citep{sun2024aligning} measure whether generated descriptions contain unsupported visual content. Attribute- and question-based benchmarks such as DLC-Bench~\citep{lian2025describe}, GAR-Bench~\citep{wang2025grasp}, Capability~\citep{liu2025goodcaption}, and CaptionQA~\citep{yang2025captionqa} further evaluate fine-grained correctness, coverage, and usefulness through localized attributes or image-grounded questions. These works motivate the view that dense captions should be evaluated at the level of visual claims. \textsc{ClaimDiff-RL} follows this direction, but uses fine-grained diagnosis inside the training reward rather than only as an evaluation protocol.


\paragraph{Reward construction for caption RL}
RL for image captioning was popularized by self-critical sequence training, which optimizes metrics such as CIDEr with policy gradients~\citep{rennie2017selfcritical}. Recent dense-caption RL methods use stronger supervision: CapRL~\citep{xing2025caprl} uses downstream QA utility as a verifiable scalar reward, while SC-Captioner~\citep{zhang2025sccaptioner} constructs decomposed rewards from parsed object, attribute, and relation sets. \textsc{ClaimDiff-RL} follows the decomposed-reward direction, but replaces fixed-schema parsing with open-vocabulary actor--reference difference verification and composes typed side-specific errors into relative or actor-only rewards.
\section{Method: Claim-Difference Rewards for Caption RL}
\label{sec:method}

\begin{figure}[t]
    \centering
    \includegraphics[width=\textwidth]{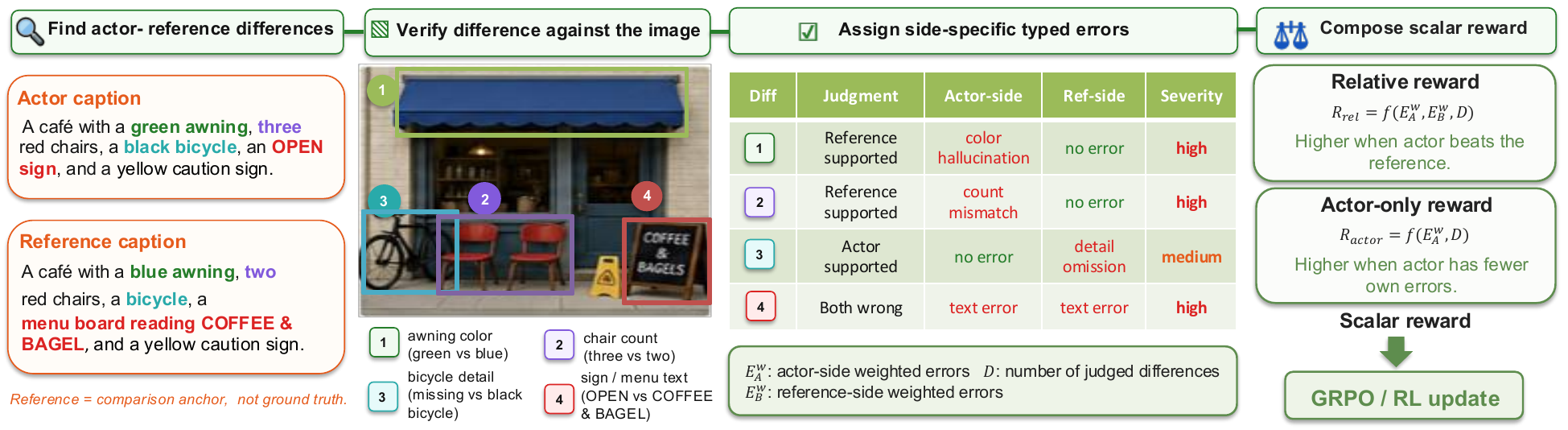}
    \caption{
    Overview of \textsc{ClaimDiff-RL}. Actor--reference differences are verified against the image to produce side-specific typed errors, which are composed into relative or actor-only scalar rewards for group-normalized RL optimization.
    }
    \label{fig:method}
\end{figure}

\textsc{ClaimDiff-RL} optimizes a scalar reward for caption RL, but obtains this scalar through decomposed judging rather than direct holistic scoring. As shown in Figure~\ref{fig:method}, given an image \(I\), an actor caption \(A \sim \pi_\theta(\cdot \mid I)\), and a reference caption \(B\), a multimodal judge first identifies concrete actor--reference visual differences, verifies each difference against the image, and assigns typed errors to the actor side and the reference side separately. The reference caption is not treated as exhaustive ground truth. It serves as a comparison anchor that proposes likely visual axes, while the image remains the verifier.

This design separates two roles that are conflated in direct scalar judging. The judge performs local verification at the level of visual claim differences, while the reward function decides how to aggregate the resulting evidence into a scalar reward. The same judge output supports two reward compositions. A relative reward compares actor-side errors against reference-side errors. An actor-only reward removes reference-side error counts from the reward and penalizes only actor-side errors on the discovered differences. Both rewards are still reference-conditioned because the reference helps define the comparison axes.

\subsection{Claim-difference judging}
\label{sec:judge}

Given \((I,A,B)\), we query a multimodal judge \(\mathcal{J}\) with a structured prompt template. The judge returns a list of \(D\) image-grounded differences,
\[
\mathcal{D}(I,A,B)=\{d_i\}_{i=1}^{D}.
\]
Each difference \(d_i\) contains a visual aspect, the actor-side claim, the reference-side claim, an image-grounded judgment, and side-specific error descriptions:
\[
d_i =
\left(
a_i,\ c_i^A,\ c_i^B,\ j_i,\ \mathbf{e}_i^A,\ \mathbf{e}_i^B
\right).
\]
Here \(a_i\) is a free-text aspect, such as \emph{awning color}, \emph{chair count}, \emph{menu text}, or \emph{background object detail}. The judgment
\[
j_i \in \{A,\ B,\ \text{both\_wrong},\ \text{both\_supported}\}
\]
indicates which side is supported by the image. The side-specific error description for caption \(X \in \{A,B\}\) is
\[
\mathbf{e}_i^X = (t_i^X,\ r_i^X,\ s_i^X),
\]
where \(t_i^X\) is an open-vocabulary error type, \(r_i^X\) is a free-text rationale, and \(s_i^X\) is an optional severity label. If caption \(X\) has no error on difference \(i\), we set \(t_i^X=\texttt{NONE}\).


The judge prompt separates difference discovery from visual verification. It first uses the textual contrast between \(A\) and \(B\) to efficiently identify candidate differences, which reduces the search space for the judge. It then verifies each candidate difference against the image, so the reference caption is not treated as ground truth. For each side, the judge assigns a specific open-vocabulary error type, preferably in a compound form such as \texttt{color\_hallucination}, \texttt{count\_mismatch}, or \texttt{detail\_omission}. The prompt also treats two common reward-hacking patterns as errors: hedging when the image supports a definite claim, and repetition that restates the same content without adding new information. 

This interface uses the reference caption as a proposal mechanism rather than as exhaustive ground truth. Textual comparison proposes candidate axes of disagreement, while image verification decides correctness. As a result, the judge can represent cases where the actor is supported, the reference is supported, both are wrong, or both are supported. The complete judge prompt and output format are provided in Appendix~\ref{app:prompts}.

\subsection{Scalar reward composition}
\label{sec:reward}

From the judge output, we compute side-specific error statistics. The unweighted error count for caption \(X \in \{A,B\}\) is
\[
E_X
=
\sum_{i=1}^{D}
\mathbf{1}\!\left[t_i^X \neq \texttt{NONE}\right].
\]
We also define a severity-weighted error count,
\[
E_X^{w}
=
\sum_{i=1}^{D}
w(s_i^X)
\cdot
\mathbf{1}\!\left[t_i^X \neq \texttt{NONE}\right],
\]
where \(w(\cdot)\) maps severity labels to non-negative weights. We use a monotone weighting scheme,
\[
w_1 \leq w_2 \leq w_3,
\]
so that more severe errors receive larger penalties. Thus, factual hallucinations or wrong counts can be penalized more strongly than minor style or wording errors. Severity can be assigned by the judge or mapped from error types. For normalization, we define
\[
D_+ = \max(D,1),
\qquad
W_{\max} = \max_s w(s).
\]
The scalar reward is then composed from the side-specific statistics \((D,E_A^w,E_B^w)\).

\paragraph{Relative ClaimDiff reward.}
The relative reward compares actor-side and reference-side weighted errors:
\[
R_{\mathrm{rel}}
=
\frac{1}{2}
-
\frac{1}{2}
\cdot
\mathrm{clamp}
\left(
\frac{E_A^w - E_B^w}{W_{\max}D_+},
-1,\ 1
\right).
\]
Thus \(R_{\mathrm{rel}}>1/2\) when the actor has fewer or less severe errors than the reference on the judge-discovered differences, and \(R_{\mathrm{rel}}<1/2\) when it has more. Because \(E_B^w\) enters the reward, this mode explicitly optimizes relative improvement against the reference. It is useful when the goal is to improve comparative quality or coverage, but it can also place stronger pressure on the actor to add specific visual claims.

\paragraph{Actor-only ClaimDiff reward.}
The actor-only reward removes reference-side error counts from the reward and penalizes only errors made by the actor on the discovered differences. For samples with at least one difference, we define the actor-side weighted error density as
\[
\rho_A =
\frac{E_A^w}{W_{\max}D}.
\]
The reward is
\[
R_{\mathrm{actor}}
=
1-\rho_A,
\qquad D>0.
\]
Thus, the actor receives reward \(1\) when it makes no actor-side errors on the discovered differences, and reward \(0\) when every discovered difference contains a maximum-severity actor-side error.

This reward is still reference-conditioned because the reference caption helps determine the comparison axes and therefore \(D\). However, unlike the relative reward, it does not use the reference-side error count \(E_B^w\). We call it actor-only because the numerator contains only actor-side errors. The actor is therefore rewarded for avoiding its own errors on the discovered visual differences, rather than for benefiting from reference-side errors. 
When the judge returns no differences, \(D=0\), assigning maximum reward can make short or non-committal captions appear artificially good. We therefore avoid a \(D=0 \Rightarrow R=1\) shortcut. For both reward compositions, zero-difference samples receive a neutral reward,
\[
R_{\mathrm{rel}} = R_{\mathrm{actor}} = r_{\mathrm{neutral}},
\qquad D=0.
\]
This prevents samples with no discovered comparison axis from becoming trivially high-reward examples while keeping the reward compatible with scalar-reward RL.

\paragraph{Ambiguity penalty.}
The actor may reduce explicit errors by using vague or disjunctive phrases, such as ``possibly'', ``might be'', or ``A or B''. The judge prompt already treats such hedging as an error when the image evidence is clear. We additionally apply a lightweight post-composition penalty to discourage systematic ambiguity:
\[
R
\leftarrow
R
\cdot
\exp
\left(
-c \cdot
\max(0,\ n_{\mathrm{amb}} - n_{\mathrm{free}})
\right).
\]
Here \(n_{\mathrm{amb}}\) is the number of detected ambiguity phrases in the actor caption, and \(n_{\mathrm{free}}\) is a length-dependent free quota. This allows occasional natural uncertainty while discouraging repeated hedging as an optimization strategy. The detection pattern and hyperparameters are specified in Appendix~\ref{app:training}.

\subsection{RL optimization}
\label{sec:rl}

After reward composition, the resulting scalar reward is used to optimize the captioning policy. Our method does not depend on a specific RL objective. In our experiments, for each image we sample multiple actor captions, score each caption with the selected \textsc{ClaimDiff-RL} reward, and use the resulting group-normalized rewards for policy optimization. Since the final reward is scalar, \textsc{ClaimDiff-RL} can be plugged into standard scalar-reward RL pipelines in the same way as a holistic judge reward. The difference is that the scalar is built from typed, image-verified claim differences rather than from a direct global score.
\section{Experiments}
\label{sec:exp}

We evaluate whether \textsc{ClaimDiff-RL} improves long-form captioning without collapsing into conservative under-captioning. Our experiments focus on three questions: whether claim-difference rewards provide a better hallucination--coverage tradeoff than holistic scalar rewards, whether the resulting captions preserve captioning capability on public benchmarks, and whether caption-side optimization maintains general VQA ability.

\subsection{Setup}
\label{sec:exp:setup}

\paragraph{Models.}
Our actor model is initialized from Qwen3-VL-32B-Instruct~\cite{Qwen3-VL} after supervised fine-tuning on long-form captions. To construct the SFT data, we randomly sample \(2\)M images from open-source image datasets, including LAION~\cite{laion5b} and DataComp-1B~\cite{datacomp}, and use Gemini-3-Pro-Preview~\citep{gemini2024team} to generate long-form reference captions. The SFT captioner is trained on these generated captions. For RL, the actor is initialized from the SFT checkpoint. The judge used in online RL is Gemini-3-Pro-preview. For \textsc{ClaimDiff-RL}, the reference caption \(B\) is generated by Gemini-3-Pro-Preview on the same image and is used as a comparison anchor rather than exhaustive ground truth.

\paragraph{Training.}
We train with GRPO~\citep{shao2024deepseekmath}. The RL training set contains \(10\)K images sampled from the SFT data pool. For each image, the policy samples 8 rollouts, each rollout is scored by the selected reward, and advantages are normalized within the rollout group. We freeze the vision tower during SFT and RL training. Unless otherwise specified, all RL variants use the same training data, actor initialization, rollout setting, and optimization recipe, so differences are attributable to the reward design. Detailed hyperparameters are provided in Appendix~\ref{app:training}.


\paragraph{Reward variants.}
We compare \textsc{ClaimDiff-RL} against holistic scalar reward baselines. The holistic-with-reference baseline asks the judge to score the actor caption given the image and a Gemini reference caption, while the holistic-no-reference baseline asks the judge to score the actor caption using only the image. Both holistic baselines directly return a scalar score on a \(0\)--\(10\) scale, which we normalize to \([0,1]\) before GRPO training. \textsc{ClaimDiff-RL} instead decomposes the actor and reference captions into claim differences, assigns side-specific typed errors according to the image, and composes scalar rewards from the resulting statistics. We evaluate both the relative reward, which compares actor-side and reference-side errors, and the actor-only reward, which penalizes actor-side errors on the discovered differences. The specific judge prompts for the holistic baselines and \textsc{ClaimDiff-RL} are shown in Appendix~\ref{app:prompts}.


\paragraph{Benchmarks.}
We evaluate three aspects of model quality: faithfulness--coverage tradeoff, public captioning capability, and general multimodal ability.

\textbf{Hallucination and missing-fact diagnostic benchmark.}
We construct a \(160\)-image human-labeled diagnostic benchmark with ground-truth captions. This benchmark is designed to distinguish two failure modes that are often conflated by scalar caption scores: unsupported visual claims and omitted salient content. Given an image \(I\), a human ground-truth caption \(R\), and a candidate caption \(C\), Gemini-3-Pro-preview performs a two-stage diagnosis. It first identifies caption-level differences between \(R\) and \(C\), including contradictions, candidate-only extra claims, and reference-only missing facts. It then verifies each contradiction or extra claim against the image. A candidate claim is counted as a hallucination only if the image contradicts it; claims that are image-supported are not penalized even when absent from \(R\). This prevents the evaluation from treating the human caption as exhaustive ground truth and allows correct extra details to receive credit. We report mean hallucination count \(\overline{\mathrm{Hall}}\), mean missing-fact count \(\overline{\mathrm{Miss}}\). The full prompt, parsing rules, and per-domain breakdowns are provided in Appendix~\ref{app:hallbench}.

\textbf{Public captioning capability.}
We evaluate fine-grained captioning ability on the captioning split of Capability~\citep{liu2025goodcaption}. We report F1 scores for sub-categories such as object category, number, color, spatial relation, scene, camera angle, OCR, and style. This benchmark tests whether reward optimization preserves the model's ability to describe detailed visual attributes beyond the hallucination diagnostic set.

\textbf{General multimodal capability.}
Finally, we evaluate whether caption-side RL affects broader visual understanding. We report VQA performance on BLINK~\citep{Fu2024BLINKML}, OCRBench-v\(2\)~\citep{Fu2024OCRBenchVA}, HRBench-\(4\)K~\citep{hrbench}, RealWorldQA~\citep{xai2024realworldqa}, and SimpleVQA~\citep{Cheng2025SimpleVQAMF}. Since these benchmarks are not optimized directly during RL, they serve as a check that the reward does not overfit to caption style at the expense of general multimodal capability.

\begin{figure}[t]
    \centering
    \includegraphics[width=\textwidth]{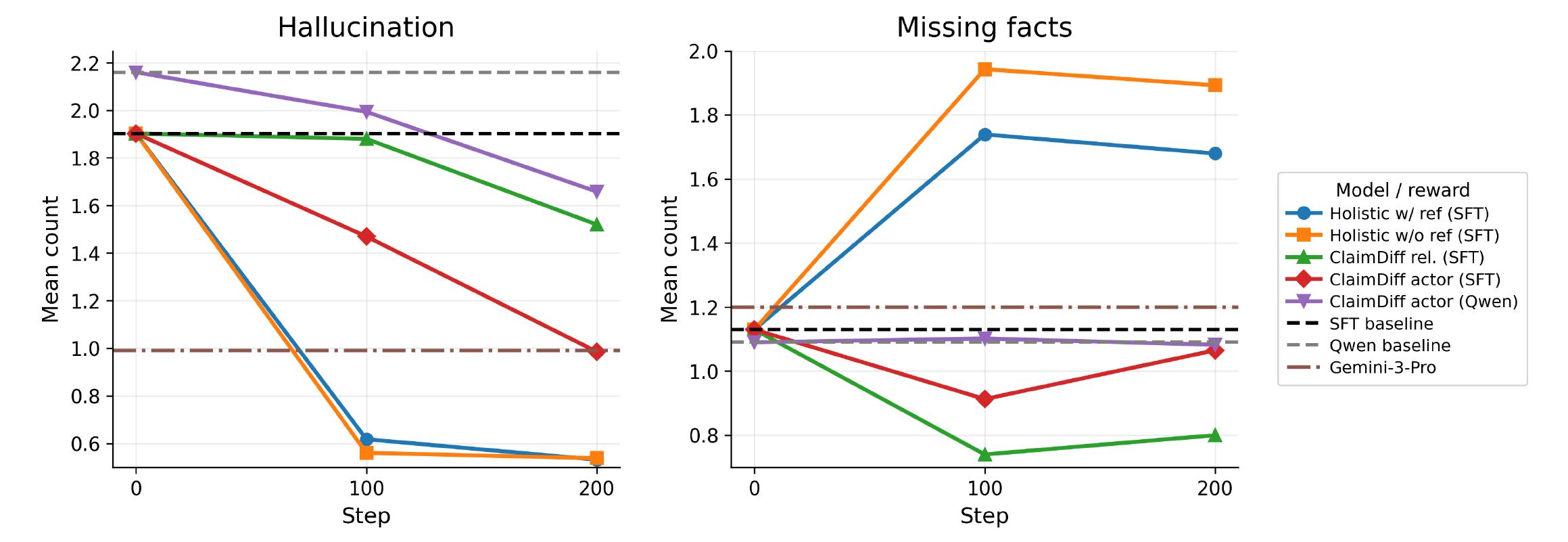}
    \caption{
    Hallucination and missing-fact trends across RL training steps. Step \(0\) denotes the corresponding initialization checkpoint. Holistic rewards reduce hallucination aggressively but increase missing facts, while \textsc{ClaimDiff-RL} exposes controllable faithfulness--coverage tradeoffs.
    }
    \label{fig:hall_miss_trends}
\end{figure}

\subsection{Results}
\label{sec:results}

\paragraph{Hallucination--missing-fact tradeoff.}
Figure~\ref{fig:hall_miss_trends} plots hallucination and missing-fact counts across RL training. Direct holistic rewards rapidly suppress hallucination, but at the cost of substantially higher missing-fact counts, especially without a reference anchor. This suggests that direct scalar rewards can be optimized by saying less. In contrast, \textsc{ClaimDiff-RL} produces more controllable operating points: the relative reward is coverage-seeking and keeps missing facts low, while the actor-only reward is more hallucination-averse and steadily reduces actor-side hallucination without a large increase in missing facts. The Qwen-initialized actor-only run follows the same trend but remains weaker than the SFT-initialized actor-only run in terms of average hallucination, indicating that a caption-specialized SFT initialization is beneficial for fine-grained caption RL.

\begin{figure}[t]
    \centering
    \includegraphics[width=\textwidth]{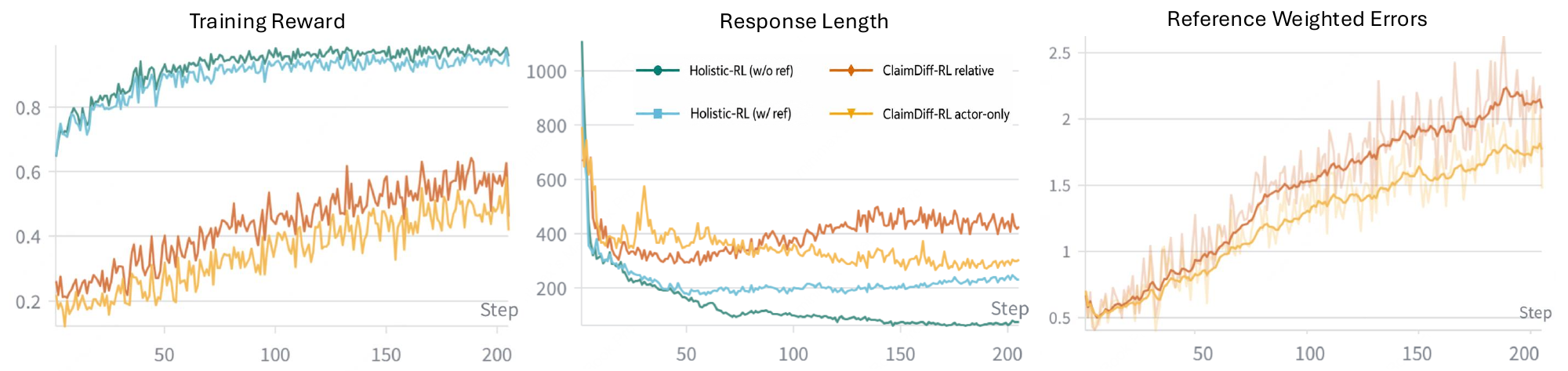}
    \caption{
Training dynamics of reward, response length, and reference-side weighted errors. 
}
    \label{fig:train_dynamics}
\end{figure}

\paragraph{Training dynamics reveal the source of under-captioning.}
Figure~\ref{fig:train_dynamics} tracks training reward, response length, and reference-side weighted errors. Holistic scalar rewards rapidly achieve high training reward while sharply reducing response length, especially without a reference anchor. This suggests that direct scalar rewards can be optimized by producing short, conservative captions that avoid risky visual claims. In contrast, \textsc{ClaimDiff-RL} reaches training reward more gradually and maintains longer captions. The relative reward keeps the longest responses, reflecting its coverage-seeking comparison against the reference, while the actor-only reward is more conservative but still avoids the severe length collapse of holistic scoring. These dynamics support Figure~\ref{fig:hall_miss_trends}: holistic rewards reduce hallucination partly through omission, whereas claim-difference rewards expose a more explicit faithfulness--coverage tradeoff.

\paragraph{Public captioning capability.}
Table~\ref{tab:capability} evaluates whether reward optimization preserves public captioning capability. Among RL-trained models, \textsc{ClaimDiff-RL} relative performs best overall, improving the average F1 from \(69.5\) for SFT to \(71.5\). More importantly, it improves several attribute-level dimensions over SFT, including object category, object number, spatial relation, scene, and camera angle, while matching SFT on OCR. Some of these gains are substantial: object number increases from \(44.1\) to \(49.8\), and spatial relation increases from \(57.9\) to \(64.2\). On selected dimensions, \textsc{ClaimDiff-RL} relative even surpasses Gemini-3-Pro-Preview, including object number, spatial relation, and scene recognition. This suggests that the relative claim-difference reward improves fine-grained, coverage-oriented captioning ability rather than merely suppressing hallucinations.

The actor-only variant is more conservative and remains close to the SFT baseline, with an average F1 of \(69.3\). It improves number, color, spatial relation, and camera angle over SFT, but drops on OCR and style. In contrast, holistic scalar rewards degrade captioning capability, especially in the no-reference setting. \textsc{Holistic-RL} (w/o ref) drops substantially on object category, object number, color, OCR, and character identification, consistent with its tendency toward conservative under-captioning.

\begin{table}[t]
\centering
\small
\caption{
Captioning capability on Capability. We report F1 for each sub-category. Best results are shown in bold and second-best results are underlined.
}
\label{tab:capability}
\resizebox{\textwidth}{!}{
\begin{tabular}{lcccccccccc}
\toprule
Model / F1 score 
& Obj. & Num. & Color & Spatial & Scene & Camera & OCR & Style & Char ID & Avg. \\
\midrule
Gemini-3-Pro-Preview 
& \textbf{83.3} & \underline{48.5} & \underline{72.8} & \underline{62.8} & \underline{80.3} & \textbf{76.6} & \textbf{96.8} & \underline{91.0} & \textbf{69.2} & \textbf{75.7} \\
Qwen3VL-32B-Instruct 
& 80.7 & 43.5 & \textbf{74.4} & 58.9 & 80.2 & 55.1 & 95.0 & \textbf{91.9} & 41.9 & 69.0 \\
SFT 
& 81.2 & 44.1 & 71.0 & 57.9 & 79.0 & 63.8 & \underline{95.4} & 88.6 & \underline{44.5} & 69.5 \\
\midrule
\textsc{Holistic-RL} (w/ ref, init: SFT) 
& 79.8 & 37.0 & 70.7 & 57.2 & 78.3 & 66.3 & 93.8 & 89.1 & 38.9 & 67.9 \\
\textsc{Holistic-RL} (w/o ref, init: SFT) 
& 74.8 & 31.1 & 67.8 & 55.6 & 78.2 & \underline{67.3} & 91.5 & 88.7 & 37.4 & 65.8 \\
\midrule
\textsc{ClaimDiff-RL} relative (init: SFT) 
& \underline{82.8} & \textbf{49.8} & 71.7 & \textbf{64.2} & \textbf{81.1} & 67.0 & \underline{95.4} & 88.1 & 43.1 & \underline{71.5} \\
\textsc{ClaimDiff-RL} actor-only (init: SFT) 
& 79.5 & 45.6 & 72.1 & 59.0 & 79.3 & 65.5 & 93.0 & 86.0 & 44.0 & 69.3 \\
\bottomrule
\end{tabular}
}
\end{table}


\paragraph{General VQA capability.}
Table~\ref{tab:vqa} evaluates whether caption-side training affects broader multimodal understanding. Supervised fine-tuning on caption data alone substantially reduces general VQA performance, dropping the average score from \(66.90\) for Qwen3VL-32B-Instruct to \(58.23\). This suggests that optimizing only for long-form caption imitation can hurt non-caption visual reasoning ability. RL partially mitigates this degradation. Among the SFT-initialized variants, \textsc{ClaimDiff-RL} relative improves the average from \(58.23\) to \(63.53\), with gains on all five VQA benchmarks, indicating that claim-difference rewards recover useful visual grounding beyond caption style. The actor-only variant is more conservative and improves the average to \(60.69\), while holistic rewards provide smaller recovery. More interestingly, applying \textsc{ClaimDiff-RL} actor-only directly from the Qwen initialization further improves the average from \(66.90\) to \(67.52\), achieving the best scores on all the benchmarks. These results suggest that claim-difference RL can act not only as a caption-quality optimizer but also as a lightweight alignment step that preserves, and in some settings improves, general multimodal capability.


\begin{table}[t]
\centering
\small
\caption{
General VQA capability across the selected benchmarks.
}
\label{tab:vqa}
\resizebox{\textwidth}{!}{
\begin{tabular}{lcccccc}
\toprule
Model / reward 
& BLINK & OCRBench-v2 & HRBench-4K & RealWorldQA & SimpleVQA & Avg. \\
\midrule
Qwen3VL-32B-Instruct 
& 64.9 & \underline{71.7} & \textbf{74.5} & \underline{70.7} & \underline{52.7} & \underline{66.90} \\
\textsc{ClaimDiff-RL} actor-only (init: Qwen) 
& \textbf{66.1} & \textbf{72.7} & \textbf{74.5} & \textbf{71.1} & \textbf{53.2} & \textbf{67.52} \\
\midrule
SFT 
& 61.6 & 68.38 & 53.5 & 62.75 & 44.91 & 58.23 \\
\textsc{Holistic-RL} (w/ ref, init: SFT) 
& \underline{65.8} & 68.7 & 64.0 & 64.1 & 44.0 & 61.32 \\
\textsc{Holistic-RL} (w/o ref, init: SFT) 
& 64.3 & 68.1 & 57.5 & 66.0 & 44.5 & 60.08 \\
\textsc{ClaimDiff-RL} relative (init: SFT) 
& 65.28 & 70.34 & \underline{66.5} & 68.76 & 46.77 & 63.53 \\
\textsc{ClaimDiff-RL} actor-only (init: SFT) 
& 63.76 & 69.24 & 62.0 & 64.18 & 44.25 & 60.69 \\
\bottomrule
\end{tabular}
}
\end{table}

\subsection{Analysis}
\label{sec:analysis}


\paragraph{Severity weighting controls the faithfulness--coverage tradeoff.}
Table~\ref{tab:severity_ablation} studies how severity weights affect the relative \textsc{ClaimDiff-RL} reward. All variants are evaluated before applying the ambiguity penalty to isolate the effect of severity weighting. We report MEDC on the \(3\)K validation set, where MEDC is the mean actor-minus-reference error count against Gemini-3-Pro-Preview captions, and report \(\overline{\mathrm{Hall}}\) and \(\overline{\mathrm{Miss}}\) on the human-labeled diagnostic benchmark. The weights \(w=(1,1,1)\) treat all error severities equally. This setting gives the lowest missing-fact count, \(0.49\), but the highest hallucination count, \(2.18\), suggesting more aggressive captioning with better coverage but more unsupported claims. Increasing \(w_2\) and \(w_3\) assigns larger penalties to more severe errors. The default setting \(w=(1,1.25,1.6)\) reduces hallucination to \(1.60\) and improves MEDC from \(0.92\) to \(0.52\), while moderately increasing missing facts to \(0.76\). A stronger setting \(w=(1,1.5,2)\) further lowers hallucination to \(1.32\), but increases missing facts to \(0.92\), indicating a shift toward more conservative captions. We therefore use \(w=(1,1.25,1.6)\) as the default operating point and provide the ambiguity-penalty ablation in Appendix~\ref{app:ambiguity_ablation}.


\begin{table}[t]
\centering
\small
\caption{
Effect of severity weighting on \textsc{ClaimDiff-RL} relative.
}
\label{tab:severity_ablation}
\begin{tabular}{lccc}
\toprule
Severity weights 
& MEDC \(\downarrow\) 
& \(\overline{\mathrm{Hall}}\) \(\downarrow\) 
& \(\overline{\mathrm{Miss}}\) \(\downarrow\) \\
\midrule
No severity weighting, \(w=(1,1,1)\)
& 0.92 & 2.18 & \textbf{0.49} \\
Default, \(w=(1,1.25,1.6)\)
& \textbf{0.52} & 1.60 & 0.76 \\
Stronger, \(w=(1,1.5,2)\)
& 0.86 & \textbf{1.32} & 0.92 \\
\bottomrule
\end{tabular}
\end{table}

\paragraph{Judge reliability and consistency.}
We further test whether our diagnostic evaluation depends on a single automatic judge. First, we manually audit Gemini-3-Pro-preview judgments with three human experts. On approximately \(100\) samples containing about \(300\) claim-level hallucination and missing-fact labels, the experts verify whether Gemini's labels are correct. Gemini reaches \(87\%\) per-claim accuracy, suggesting that the typed claim-level judgments are reliable enough for aggregate diagnostic evaluation. Second, we compare Gemini-3-Pro-preview with GPT-5.2~\citep{singh2026openaigpt5card} under the same reference-conditioned evaluation protocol used in our main diagnostic benchmark. On the same three model families, we compute per-sample Spearman correlations for hallucination and missing-fact counts. The two judges show positive agreement, with \(\rho=0.537\) for hallucination and \(\rho=0.334\) for missing facts. Hallucination is more consistently judged than missing facts, suggesting that unsupported claims are easier to verify than omitted content. These results indicate that our main hallucination--coverage conclusions are not an artifact of a single automatic judge. More analysis about judge reliability and consistency is presented in Appendix~\ref{app:judge_consistency}.


\paragraph{Reference-conditioned vs. no-reference diagnosis.}
Finally, we analyze the role of the reference caption during diagnostic evaluation. We use the \textsc{ClaimDiff-RL} relative checkpoint and evaluate it on the same \(160\)-image human-labeled diagnostic benchmark. Table~\ref{tab:ref_noref_eval} compares Gemini-based judging with and without the human reference caption. Removing the reference reduces the number of detected errors: average hallucinations decrease from \(1.52\) to \(0.97\), and average missing facts decrease from \(0.80\) to \(0.44\). This does not necessarily mean that the captions are better under no-reference judging. Instead, it indicates that without a reference anchor, the judge identifies fewer comparison axes, especially for omissions. This supports our design choice: the reference caption is useful not as ground truth, but as a proposal mechanism for salient visual claims that should then be verified against the image.

\begin{table}[t]
\centering
\small
\caption{
Reference-conditioned vs. no-reference diagnostic on the human-labeled benchmark. 
}
\label{tab:ref_noref_eval}
\begin{tabular}{lccc}
\toprule
Metric & Ref-conditioned & No-reference & Difference \\
\midrule
Avg. hallucination / sample \(\downarrow\) & 1.52 & 0.97 & \(-0.55\) \\
Avg. missing fact / sample \(\downarrow\) & 0.80 & 0.44 & \(-0.36\) \\
Samples with hallucination \(\downarrow\) & 111 & 77 & \(-34\) \\
\bottomrule
\end{tabular}
\end{table}

\section{Conclusion}
\label{sec:conclusion}


We introduced \textsc{ClaimDiff-RL}, a fine-grained reward framework for long-form image captioning that obtains scalar rewards from image-verified actor--reference claim differences rather than direct holistic caption scores. By separating hallucinated claims, missing facts, and correct extra details before reward composition, \textsc{ClaimDiff-RL} makes the faithfulness--coverage tradeoff explicit and diagnosable. Empirically, we find that holistic scalar rewards can reduce hallucination by encouraging conservative under-captioning, while \textsc{ClaimDiff-RL} provides more controllable operating points: the relative reward improves coverage-oriented captioning, and the actor-only reward better suppresses actor-side hallucination. Across diagnostic, public captioning, and VQA evaluations, claim-difference rewards preserve or improve downstream capability compared with SFT and holistic scalar optimization. Notably, on Capability, \textsc{ClaimDiff-RL} surpasses Gemini-3-Pro-Preview on selected fine-grained dimensions such as object counting, spatial relations, and scene recognition, suggesting that typed, verifiable claim differences are an effective reward unit for diagnosable multimodal RL.


\bibliographystyle{plain}
\bibliography{ref}

\appendix



\newpage
\section{Limitations}
\label{app:limitations}

\paragraph{Dependence on strong multimodal judges.}
\textsc{ClaimDiff-RL} relies on a strong multimodal judge to identify actor--reference differences, verify them against the image, and assign typed errors. Although our human audit and GPT--Gemini consistency analysis suggest that the automatic judgments are reliable at the aggregate level, individual claim-level judgments can still be noisy. Errors may arise from difficult visual evidence, OCR ambiguity, small objects, subjective style descriptions, or cases where the image does not clearly support either caption. As a result, \textsc{ClaimDiff-RL} should be interpreted as a framework for scalable fine-grained reward construction rather than a replacement for human evaluation.

\paragraph{Reference-conditioned comparison is not reference-free.}
The reference caption is used as a comparison anchor to propose candidate visual axes. This reduces the judge's search space and improves consistency, but it also means that the discovered differences are influenced by the reference caption's coverage and style. If the reference omits an important visual aspect, the judge may be less likely to evaluate that aspect. Conversely, if the reference contains unusual or noisy details, the actor may be compared along less useful axes. We mitigate this by verifying every difference against the image and by not treating the reference as exhaustive ground truth, but the reward remains reference-conditioned.

\paragraph{Potential reward hacking.}
Decomposed rewards make some failure modes easier to diagnose, but they do not eliminate reward hacking. Models may still learn to exploit judge preferences, repeat safe details, hedge in ways not captured by our ambiguity parser, or optimize for the style of the judge prompt. Our ambiguity penalty and zero-difference handling reduce some observed shortcuts, but more robust defenses, such as judge ensembles, adversarial audits, stronger uncertainty handling, and periodic human evaluation, remain important.

\paragraph{Evaluation scope.}
Our diagnostic benchmark contains \(160\) human-labeled images and is designed to measure hallucination and missing facts in long-form captions. While it is useful for controlled analysis, it is not exhaustive. The benchmark may not cover all domains, rare visual concepts, complex documents, highly specialized OCR, medical or scientific imagery, or culturally specific entities. We therefore complement it with public captioning and VQA benchmarks, but broader evaluation across more domains and languages is needed.

\paragraph{Computational cost.}
\textsc{ClaimDiff-RL} is more expensive than direct scalar scoring because the judge must enumerate differences and provide side-specific typed errors. This additional cost is useful for diagnosis and reward composition, but may limit scaling if applied to very large training sets or many rollout samples. Future work could reduce cost by caching reference-side analyses, using smaller specialized judges, distilling the judge into a verifier model, or applying claim-difference judging selectively to uncertain or high-value samples.

\section{Broader Impact}
\label{app:broader_impact}

\textsc{ClaimDiff-RL} aims to make long-form image captioning more reliable by decomposing caption quality into image-grounded claim differences, which can benefit accessibility, image retrieval, education, and dataset curation. However, more detailed captions may also make incorrect outputs appear more persuasive, especially in high-stakes domains such as medical, legal, or safety-critical settings. The framework still depends on automatic judges and reference captions, which may contain biases, omissions, or judge-specific preferences. Although our method helps diagnose hallucination, missing facts, ambiguity, and reward-hacking behaviors, it does not guarantee factual correctness. Responsible use should include human evaluation, diverse benchmarks, and monitoring for systematic failures.

\section{Training and Implementation Details}
\label{app:training}


\paragraph{RL training.}
We train with GRPO~\citep{shao2024deepseekmath}. The RL training set
contains $10$K images sampled from the SFT data pool. For RL rollouts,
we use a simple captioning instruction, \emph{``Please describe this
image in detail.''} We use 4 $\times$ 8 H100 GPUs with a global batch size of $32$.
Each prompt receives $8$ rollouts, and advantages are computed with a
mean baseline within each rollout group. The learning rate is
$1\times 10^{-6}$ with cosine decay. The vision tower is frozen
throughout training, while the LLM and projection layers are updated.

Unless otherwise specified, all reward variants---\textsc{ClaimDiff-RL}
relative, \textsc{ClaimDiff-RL} actor-only, \textsc{Holistic-RL}
(w/ ref), and \textsc{Holistic-RL} (w/o ref)---use the same training
data, actor initialization, rollout setting, and optimization recipe.
We select the checkpoint at step $200$ for all variants, because we
observe performance degradation after further training. This controlled
setup makes the reward design the primary experimental variable.

\subsection{Ambiguity penalty implementation}
\label{app:amb-impl}

The ambiguity penalty
$R \leftarrow R \cdot \exp\!\big(-c\cdot\max(0,\,n_{\mathrm{amb}}-n_{\mathrm{free}})\big)$
of the main text is applied \emph{after} reward composition and
before advantage normalization, so that it shifts the reward of an
individual rollout without altering the structure of the
\textsc{ClaimDiff-RL} reward family. Three pieces need to be
specified: the ambiguity-token detector, the length-dependent free
quota $n_{\mathrm{free}}$, and the decay coefficient $c$.

\paragraph{Ambiguity token detection.}
$n_{\mathrm{amb}}$ is the number of regex matches in the actor
caption $A$ against a fixed list of hedge / disjunction patterns.
The list groups four kinds of constructions:

\begin{itemize}\itemsep0pt
\item \emph{Epistemic hedges}: \texttt{possibly}, \texttt{probably},
  \texttt{perhaps}, \texttt{maybe}, \texttt{seems}, \texttt{appears},
  \texttt{looks like}, \texttt{might (be)}, \texttt{may (be)},
  \texttt{could (be)}.
\item \emph{Approximators}: \texttt{about}, \texttt{around},
  \texttt{approximately}, \texttt{roughly}, \texttt{some kind of},
  \texttt{a sort of}.
\item \emph{Disjunctions over visual claims}: a token-window match
  for the patterns \texttt{X or Y}, \texttt{either X or Y},
  \texttt{X / Y}, restricted to head nouns / colour words /
  numerals so that natural language ``or'' (e.g.,
  \emph{``a~man or a~woman''} when both are equally plausible) is
  caught while non-visual ``or'' (e.g.,
  \emph{``a poster advertising a film or concert''}) is filtered out
  by a small POS / lexicon allow-list.
\item \emph{Negated certainty}: \texttt{not clearly},
  \texttt{not entirely sure}, \texttt{hard to tell},
  \texttt{difficult to say}.
\end{itemize}

Matches are case-insensitive and counted with multiplicity; nested
matches inside a longer phrase are collapsed to a single match. The
exact token list and regex are released in the supplementary code.

\paragraph{Length-dependent free quota.}
A long, content-rich caption naturally accommodates more uncertain
phrases than a short one, and we want to penalise only \emph{systematic}
hedging, not occasional natural uncertainty. We therefore allow a
length-proportional free budget
\[
  n_{\mathrm{free}}
  \;=\;
  \big\lfloor \mathrm{len}(A) / \tau \big\rfloor,
  \qquad \tau = 90 \text{ words},
\]
i.e.\ one free hedge per $90$ words of the actor caption. With our
typical caption length ($150$--$300$ words) this gives a quota of
$1$--$3$ hedges per caption before the penalty starts to fire. Only
the excess $\max(0,\,n_{\mathrm{amb}} - n_{\mathrm{free}})$ enters
the exponent.

\paragraph{Decay coefficient.}
The decay coefficient is $c = 0.1$ throughout the paper. Concretely,
each excess hedge multiplies the reward by $e^{-0.1} \approx 0.905$,
two excess hedges by $\approx 0.819$, five excess hedges by
$\approx 0.607$. The penalty is therefore mild for small overshoots
but compounds quickly enough to dominate the gradient when the
actor begins to hedge most of its claims. We did not separately tune
$c$; the value was chosen so that a single excess hedge produces a
visible but recoverable reward decrement.

\paragraph{What the penalty does and does not do.}
By construction the penalty is monotonic in $n_{\mathrm{amb}}$ and
multiplicative in the reward, so it never inverts the sign of the
gradient on \textsc{ClaimDiff-RL}'s claim-level statistics; it only
re-weights the rollouts within a group. This is what allows us to
report it as a \emph{safeguard} on the relative reward branch in
\S\ref{sec:analysis} rather than as a separate reward objective: the
recipe still optimizes the same claim-difference reward, but rollouts
that try to game the judge through systematic hedging receive a
multiplicative discount that shifts the GRPO advantage away from
them.

\section{Reward Prompts}
\label{app:prompts}

This section provides the complete prompt templates used for all
reward variants in our experiments: the \textsc{ClaimDiff-RL}
claim-difference prompt (\S\ref{app:prompt:claimdiff}), and the two
holistic scalar baselines (\S\ref{app:prompt:holistic}).

\subsection{ClaimDiff-RL reward prompt}
\label{app:prompt:claimdiff}

The claim-difference reward judge receives the image, the actor
caption (Caption~A), and the Gemini-3-Pro reference caption
(Caption~B). The prompt instructs the judge to enumerate concrete
differences, assign per-side typed errors with model-judged severity,
and select an overall winner. The complete prompt is reproduced
verbatim below.

\begin{small}
\begin{verbatim}
You are comparing two image captions based strictly on the image.

Your task:
- Identify one or more concrete differences between Caption A
  and Caption B.
- For each difference, judge which caption is supported by the
  image.
- Describe errors for each caption if present.
- Assign an overall winner.
- For each caption error, assign a SEVERITY_LEVEL based on the
  ERROR_TYPE (model-judged, not post-mapped).

SEVERITY LEVEL GUIDELINES:
- Priority order: hallucination/incorrect claim > detail
  omission/incomplete info > style/aesthetic error.
- Severity 3 (major): The caption asserts something false or
  nonexistent, wrong identity, wrong count, wrong relation,
  or fabricated content.
- Severity 2 (medium): The caption misses, truncates, or
  incompletely describes factual details that are present
  but secondary.
- Severity 1 (minor): The caption mainly errs in style,
  aesthetic impression, tone, or subjective interpretation
  without factual contradictions.
- If ERROR_TYPE is NONE, set SEVERITY_LEVEL: NA.

AMBIGUITY HANDLING RULES:
- If the image clearly supports a specific visual attribute
  (such as a definite color, object identity, spatial relation,
  lighting condition, or camera angle), then an ambiguous or
  disjunctive claim (e.g., "A or B", "possibly A", "might be A")
  should be treated as an error.
- In such cases, prefer assigning an appropriate ERROR_TYPE
  (e.g., color_hallucination, identity_hallucination, or
  <dimension>_misinterpretation) rather than considering the
  claim acceptable.
- If the image itself is genuinely ambiguous or lacks sufficient
  visual evidence, ambiguity may be acceptable and should not
  be penalized.
- Do not reward ambiguity as a safe alternative when clear
  visual evidence is present.

ERROR TYPE GUIDELINES:
- ERROR_TYPE should be specific and descriptive, not generic.
- Use a compound form when possible:
  <dimension>_<error_nature>.
- Avoid vague terms like "hallucination" or "omission" alone.
- Prefer fine-grained types such as:
  color_hallucination, layout_omission, text_omission,
  text_truncation_error, count_mismatch,
  spatial_relation_error, style_misinterpretation,
  identity_hallucination, context_overreach,
  lighting_misinterpretation,
  camera_angle_misinterpretation, etc.
- If multiple issues exist, choose the most dominant one.
- If the image clearly supports a specific visual attribute
  (such as a definite color, object identity, spatial relation,
  lighting condition, recognizable text, or camera angle), then
  an ambiguous or disjunctive claim (e.g., "A or B",
  "possibly A", "might be A", several Chinese characters, etc.)
  should be treated as an error.

IMPORTANT FORMAT RULES (CRITICAL - MUST FOLLOW EXACTLY):
- Use EXACTLY the format below.
- Use the exact headers: [DIFFERENCE 1], ASPECT:,
  CAPTION_A_CLAIM:, CAPTION_B_CLAIM:, JUDGMENT:,
  CAPTION_A_ERROR:, ERROR_TYPE:, ERROR_DETAIL:,
  SEVERITY_LEVEL:, CAPTION_B_ERROR:, [OVERALL_WINNER]
- Do NOT use markdown, code blocks, or any other formatting.
- If a caption has no error, use:
  ERROR_TYPE: NONE
  ERROR_DETAIL: No error.
  SEVERITY_LEVEL: NA
- You MUST include the [OVERALL_WINNER] section at the end.

OUTPUT FORMAT:

[DIFFERENCE 1]
ASPECT:
CAPTION_A_CLAIM:
CAPTION_B_CLAIM:
JUDGMENT: <A|B|both_wrong|both_supported>

CAPTION_A_ERROR:
ERROR_TYPE:
ERROR_DETAIL:
SEVERITY_LEVEL: <1|2|3|NA>

CAPTION_B_ERROR:
ERROR_TYPE:
ERROR_DETAIL:
SEVERITY_LEVEL: <1|2|3|NA>

[OVERALL_WINNER]
<A|B|Tie>

Caption A:
{caption_a}

Caption B:
{caption_b}
\end{verbatim}
\end{small}

\subsection{Holistic scalar reward prompts}
\label{app:prompt:holistic}

The holistic baselines ask the judge to score the actor caption
on a $0$--$10$ scale. We evaluate two variants: with reference
and without reference. Both return a single scalar; the reward
is $\texttt{SCORE}/10$.

\paragraph{Holistic with reference.}

\begin{small}
\begin{verbatim}
You are an expert evaluator for long-form image captions.

Given an image, an actor caption, and a reference caption,
evaluate the actor caption with respect to the image.

The reference caption is provided only as a helpful comparison
anchor. It may be incomplete or contain mistakes. Do not treat
it as exhaustive ground truth.

Score the actor caption from 0 to 10 based on:
1. Visual factual correctness.
2. Coverage of salient image content.
3. Correct attributes, counts, spatial relations, OCR/text,
   and identities.
4. Avoidance of hallucinated objects, attributes, or relations.
5. Clarity and specificity without unnecessary ambiguity or
   repetition.

Important rules:
- A correct detail in the actor caption should not be penalized
  merely because it is absent from the reference.
- A detail in the reference should not be rewarded unless it is
  supported by the image.
- Penalize hallucination more than omission.
- Penalize strategic hedging when the image evidence is clear.
- Do not reward length by itself.
- Do not reward flowery style by itself.

Actor caption:
{actor_caption}

Reference caption:
{reference_caption}

Return exactly this format:
SCORE: <integer from 0 to 10>
RATIONALE: <one short sentence>
\end{verbatim}
\end{small}

\paragraph{Holistic without reference.}

\begin{small}
\begin{verbatim}
You are an expert evaluator for long-form image captions.

Given an image and an actor caption, evaluate the actor caption
with respect to the image.

Score the actor caption from 0 to 10 based on:
1. Visual factual correctness.
2. Coverage of salient image content.
3. Correct attributes, counts, spatial relations, OCR/text,
   and identities.
4. Avoidance of hallucinated objects, attributes, or relations.
5. Clarity and specificity without unnecessary ambiguity or
   repetition.
   

Important rules:
- Penalize hallucination more than omission.
- Penalize strategic hedging when the image evidence is clear.
- Do not reward length by itself.
- Do not reward flowery style by itself.

Actor caption:
{actor_caption}

Return exactly this format:
SCORE: <integer from 0 to 10>
RATIONALE: <one short sentence>
\end{verbatim}
\end{small}

\section{Hallucination and Missing-Fact Diagnostic Benchmark}
\label{app:hallbench}

\subsection{Image set and references}

We use a fixed $N{=}160$-image diagnostic benchmark with
human-written ground-truth captions. The set is fixed before model
evaluation and used identically across all checkpoints reported in
the paper. We select the 160 images from publicly available benchmarks and balance the domain on purpose. Therefore, differences in
Figure~\ref{fig:hall_miss_trends} and the corresponding diagnostic
metrics are attributable to model behaviour rather than
evaluation-set sampling.

Each image $I$ is paired with a human-written ground-truth caption
$R$. A model under evaluation generates a candidate caption $C$.
The benchmark measures two complementary failure modes: hallucinated
visual claims, where $C$ states content contradicted by the image,
and missing facts, where $C$ omits salient content described in $R$.
This makes the benchmark suitable for analysing whether reward
optimisation reduces hallucination by improving visual grounding or
merely by producing more conservative captions.

\subsection{Candidate caption generation prompt}
\label{app:caption_prompt}

For each model checkpoint evaluated on the diagnostic benchmark, we use the same caption-generation prompt to obtain the candidate caption \(C\). The prompt asks the model to produce both an overview caption and a detailed caption in English. We use the generated \texttt{Detailed\_Description} as the candidate caption \(C\) for hallucination and missing-fact evaluation. The prompt emphasizes visually observable content, including objects, counts, colors, spatial relations, OCR/text, scene details, style, camera angle, image quality, and other salient visual elements, while discouraging unsupported speculation. The exact prompt is shown below.

\textbf{Caption generation prompt.}
\begin{small}
\begin{verbatim}
You are a professional image text prompt generator. 

Your task is to generate the overview and detail image 
text prompts for a given image. Based on the following 
rules, generate the text prompt for the image generation 
task in English.

Image text prompts can refer to the following aspects to 
ensure accuracy and conciseness: subject, quantity, subject 
color, composition, spatial relationships between multiple 
subjects, colors of multiple entities, perspective relationships, 
text information, photography style, art style, shooting angle. 
Accurately identify the subject and quantity of the image. 
If the subject is a landmark, specific product, celebrity, or 
other key information, inject these key names into the prompt. 
Choose specific output aspects based on actual needs, 
not necessarily including all aspects.

Focus solely on visually observable content. Describe the main subject, 
subject position and interaction, number of subjects, 
foreground and background scene information, text content and text 
layout when present, visual style, color style, photography style, 
lens information, camera angle, lighting, image quality, borders, 
watermarks, logos, screenshots, composite images, and other salient 
visual elements. If metadata is available, use it only when it helps 
identify visually relevant information, and do not reveal that the 
information comes from metadata. Do not speculate about deeper 
meanings or intents. Do not translate text appearing in the image; 
preserve the original text form.

Image text prompts should not contain descriptions unrelated to the 
image. When describing an image, start from the image itself. 
If the image does not contain a certain type of information, 
such as text, there is no need to describe that item. Do not answer 
in chunks. The output should be a complete and fluent natural-language 
paragraph. The sentence format should be diverse.

Detailed_Description: According to the above rules, provide a 
relatively detailed description of the basic information of the image. 
The language should be natural and fluent, controlled within 200 to 
300 words, expressed as a complete paragraph without dividing into 
paragraphs.

Overview_Description: Summarize the main information of the image 
in one sentence, clearly and concisely stating the basic information 
of the subject, quantity information, interaction relationships, 
background information, art style, image color style, image lens 
information, image photography angle, and photography style. If 
there is text in the image, include it in the description.

Directly output the content of the image. Do not start with phrases 
such as ``The image features'', ``The image captures'', or ``The image 
prompt features''. Please ensure the analysis is comprehensive and detailed. 
Do not return explanatory content. Directly return image-related 
information with Overview_Description and Detailed_Description. 
The results should be presented in English.

Return in the following YAML structure:


yaml
Overview_Description: |
   The overview caption.
Detailed_Description: |
   The detailed caption.
\end{verbatim}
\end{small}

\subsection{Vision judge and prompt schema}

The diagnostic judge is \textsc{Gemini-3-Pro-preview} with
deterministic decoding. We use an enforced JSON
\texttt{response\_schema}; rare malformed responses are re-queried up
to three times and otherwise excluded from the aggregate. The judge
takes three inputs: the image $I$, the human ground-truth caption
$R$, and the candidate caption $C$. It then performs a two-stage
evaluation.

\begin{enumerate}\itemsep0pt
\item \textbf{Difference detection.}
The judge compares $R$ and $C$ and emits each detected difference as
one of three types: \texttt{contradiction}, \texttt{extra\_info}, or
\texttt{missing\_fact}. A \texttt{contradiction} is a candidate claim
that conflicts with the reference. An \texttt{extra\_info} item is a
candidate claim not mentioned in the reference. A
\texttt{missing\_fact} is a fact in the reference that is absent from
the candidate.

\item \textbf{Image verification.}
For each \texttt{contradiction} or \texttt{extra\_info} item, the
judge verifies the candidate-side claim against the image and assigns
one of three verification labels: \texttt{verified},
\texttt{false}, or \texttt{ambiguous}. The label \texttt{verified}
means the image supports the candidate claim, \texttt{false} means
the image contradicts it, and \texttt{ambiguous} means the image is
insufficiently informative. \texttt{missing\_fact} items are counted
as omissions rather than hallucinations.
\end{enumerate}

This two-stage design prevents the human caption from being treated
as exhaustive ground truth. A candidate can add details that are
absent from $R$ without being penalised, as long as those details are
supported by the image. Conversely, unsupported candidate-side
details are counted as hallucinations even if they appear fluent or
plausible. Each difference is additionally tagged as either
\texttt{natural} or \texttt{design}, enabling per-domain analysis
when needed.

The full prompt text and JSON schema are released with the
supplementary code. Below we reproduce both the
reference-conditioned and no-reference diagnostic prompts verbatim.

\paragraph{Reference-conditioned diagnostic prompt.}

\begin{small}
\begin{verbatim}
You are an expert at detecting hallucinations in image captions
by comparing them with the actual image content and a ground
truth caption.

Your task has two steps:

STEP 1: Compare GT caption with Predicted caption to identify
differences. Differences can be of three types:
  * CONTRADICTION: Predicted caption says something that
    contradicts GT
  * EXTRA_INFO: Predicted caption includes information
    not in GT
  * MISSING_FACT: GT includes information that is missing
    from predicted caption

STEP 2: For EACH difference found, verify the PREDICTED
CAPTION's claim against the IMAGE:
- For CONTRADICTION:
  * VERIFIED: Predicted caption's claim IS confirmed in the
    image -> is_hallucination=false
  * FALSE: Predicted caption's claim is NOT in the image
    -> is_hallucination=true
  * AMBIGUOUS: Cannot determine from image
    -> is_hallucination=false
- For EXTRA_INFO:
  * VERIFIED: Predicted caption claims something IS in the
    image -> is_hallucination=false
  * FALSE: Predicted caption claims something NOT visible
    in the image -> is_hallucination=true
  * AMBIGUOUS: Cannot determine from image
    -> is_hallucination=false
- For MISSING_FACT:
  * This is NOT a hallucination (it's an omission)
  * Set verification="missing" and is_hallucination=false

Strict mapping rules for is_hallucination:
- type="contradiction" + verification="verified"
  -> is_hallucination=false
- type="contradiction" + verification="false"
  -> is_hallucination=true
- type="contradiction" + verification="ambiguous"
  -> is_hallucination=false
- type="extra_info" + verification="false"
  -> is_hallucination=true
- type="extra_info" + verification="verified"
  -> is_hallucination=false
- type="extra_info" + verification="ambiguous"
  -> is_hallucination=false
- type="missing_fact" -> is_hallucination=false

Category Classification:
- NATURAL: Hallucinations about natural/physical objects,
  scenes, people, animals, actions, poses, clothing,
  physical attributes, spatial relationships.
- DESIGN: Hallucinations about text content, typography,
  layout, design elements, UI elements, logos, brand names.

Guidelines:
- Be strict about contradictions
- Be lenient about description variations
- Extra details that ARE in the image should NOT be marked
  as hallucinations (even if not in GT)
- If uncertain, mark as AMBIGUOUS (not hallucination)

Ground Truth Caption:
{gt_caption}

Predicted Caption:
{pred_caption}

Return your answer in valid JSON format with this structure:
{
  "has_hallucination": true or false,
  "differences": [
    {
      "type": "contradiction" or "extra_info"
             or "missing_fact",
      "content": "...",
      "category": "natural" or "design",
      "verification": "verified" or "false"
                     or "ambiguous" or "missing",
      "reason": "...",
      "is_hallucination": true or false
    }
  ]
}
\end{verbatim}
\end{small}

\paragraph{No-reference diagnostic prompt.}

\begin{small}
\begin{verbatim}
You are an expert evaluator for long-form image caption
factuality and coverage.

You are given an image and a model-generated caption.
Evaluate the caption without using any reference caption.

Your evaluation has two parts.

Part A: Hallucination detection.
Split the caption into atomic visual claims. For each claim,
verify whether it is supported by the image. Mark unsupported
claims as HALLUCINATION. Mark unclear cases as UNCERTAIN.

Part B: Missing-fact detection.
Create a concise checklist of salient visual facts that a good
long-form caption should mention. Include only visually clear
and important facts. Then check whether the model caption
covers each fact. Mark uncovered salient facts as MISSING.

Important rules:
1. Only judge visually grounded content.
2. Do not penalize missing minor background details.
3. Do not hallucinate facts in the checklist.
4. For hallucination, judge only claims made by the caption.
5. For missing facts, judge only important image facts.
6. If visual evidence is ambiguous, mark UNCERTAIN.
7. Keep all claims atomic.

Model Caption:
{pred_caption}

Return your answer in valid JSON format with this structure:
{
  "claims": [
    {
      "claim": "<atomic claim from caption>",
      "aspect": "<object | attribute | count | spatial
                | action | text_ocr | identity | scene
                | style | other>",
      "judgment": "<SUPPORTED | HALLUCINATION | UNCERTAIN>",
      "evidence": "<brief visual evidence>"
    }
  ],
  "missing_facts": [
    {
      "fact": "<salient visual fact from image>",
      "aspect": "<...>",
      "coverage": "<COVERED | PARTIALLY_COVERED | MISSING
                  | UNCERTAIN>",
      "caption_evidence": "<caption phrase or NONE>",
      "reason": "<brief reason>"
    }
  ]
}
\end{verbatim}
\end{small}

\subsection{Deterministic mapping to hallucination and missing facts}

The final hallucination label is determined by a deterministic
post-processing rule applied to the structured judge output. This
avoids relying on the judge's free-text rationale.

\begin{center}
\small
\begin{tabular}{llc}
\toprule
\textbf{Difference type} & \textbf{Image verification} & \textbf{Counted as hallucination} \\
\midrule
\texttt{contradiction} & \texttt{false}     & \textbf{yes} \\
\texttt{contradiction} & \texttt{verified}  & no \\
\texttt{contradiction} & \texttt{ambiguous} & no \\
\texttt{extra\_info}   & \texttt{false}     & \textbf{yes} \\
\texttt{extra\_info}   & \texttt{verified}  & no \\
\texttt{extra\_info}   & \texttt{ambiguous} & no \\
\texttt{missing\_fact} & any                & no, counted as missing fact \\
\bottomrule
\end{tabular}
\end{center}

This mapping has two important consequences. First, image-supported
candidate claims are not counted as hallucinations even when they are
absent from the human reference. Second, ambiguous claims are
conservatively treated as non-hallucinations. Therefore, the
hallucination count emphasises visually contradicted claims and should
be interpreted as a conservative estimate of unsupported content.

\subsection{Metrics}

For image $I_i$, let $H_i$ denote the set of candidate-side claims
counted as hallucinations and $M_i$ denote the set of
\texttt{missing\_fact} items. We define
\[
\mathrm{Hall}_i = |H_i|,
\qquad
\mathrm{Miss}_i = |M_i|.
\]

The benchmark-level metrics are simple means over the $N{=}160$
images:
\[
\overline{\mathrm{Hall}}
=
\frac{1}{N}\sum_{i=1}^{N}\mathrm{Hall}_i,
\qquad
\overline{\mathrm{Miss}}
=
\frac{1}{N}\sum_{i=1}^{N}\mathrm{Miss}_i.
\]

The main text focuses on $\overline{\mathrm{Hall}}$ and
$\overline{\mathrm{Miss}}$, since their joint behaviour directly
reveals the faithfulness--coverage tradeoff. In particular, a model
can reduce $\overline{\mathrm{Hall}}$ by saying less, but this
typically increases $\overline{\mathrm{Miss}}$. This is why
Figure~\ref{fig:hall_miss_trends} reports both quantities across
training steps.

We deliberately do not length-normalise $\mathrm{Hall}_i$. A verbose
caption that introduces more unsupported claims should be penalised,
even if the rate of errors per token is low. At the same time,
$\overline{\mathrm{Miss}}$ prevents overly short or conservative
captions from appearing artificially strong.

\subsection{Why this benchmark is needed}

This benchmark complements public captioning and VQA evaluations.
Public captioning benchmarks measure whether models retain
fine-grained captioning capability across categories such as object,
number, colour, spatial relation, scene, camera angle, OCR, and
style. VQA benchmarks measure broader multimodal understanding.
However, neither directly separates two caption-specific failure
modes: hallucinating unsupported details and omitting salient details.

The diagnostic benchmark is designed to isolate this tradeoff. It
shows that direct holistic scalar rewards can reduce hallucination
aggressively, but often increase missing facts, indicating
conservative under-captioning. In contrast, \textsc{ClaimDiff-RL}
exposes more controllable operating points. The relative reward is
more coverage-seeking and tends to keep missing facts low, while the
actor-only reward is more hallucination-averse and reduces actor-side
hallucination without a large missing-fact increase. These trends are
summarised in Figure~\ref{fig:hall_miss_trends}.

\section{Judge Reliability and Consistency}
\label{app:judge_consistency}

Our main diagnostic benchmark relies on automatic claim-level
judgments to measure hallucinations and missing facts. We therefore
conduct additional checks to evaluate whether the conclusions are
robust to judge choice and whether the automatic judgments are
reliable at the claim level.

\paragraph{Human expert audit of Gemini judgments.}
We manually audit Gemini-3-Pro-preview judgments with three human
experts. The audit covers approximately $100$ samples and $300$
claim-level labels, including both hallucination and missing-fact
annotations. For each claim-level label produced by Gemini, human
experts verify whether the label is correct with respect to the
image and the provided ground-truth caption. Gemini reaches $87\%$
per-claim accuracy under this audit. This provides evidence that the
typed claim-level judgments are sufficiently reliable for aggregate
diagnostic analysis, while still leaving room for noise at the
individual-claim level.

\paragraph{GPT--Gemini consistency.}
We further compare Gemini-3-Pro-preview with GPT-5.2 on the same
three model families: the SFT baseline, \textsc{ClaimDiff-RL}
relative, and \textsc{ClaimDiff-RL} actor-only. For each sample, both
judges produce hallucination and missing-fact counts. We then compute
Spearman correlations between the two judges at the per-sample level.
Table~\ref{tab:app_judge_consistency_overall} summarises the overall
correlations under reference-conditioned and no-reference diagnostic
judging.

Reference-conditioned judging yields stronger agreement than
no-reference judging. With a reference caption, the hallucination
correlation is $0.537$, compared with $0.377$ without reference.
The missing-fact correlation also improves from $0.284$ to $0.334$.
This supports our use of the reference caption as a comparison anchor:
it provides shared visual axes for different judges, while the final
correctness decision is still made against the image. Hallucination is
consistently more correlated than missing facts, suggesting that
unsupported claims are easier to verify than omissions.

\begin{table}[h]
\centering
\small
\caption{
Overall Spearman correlation between GPT-5.2 and Gemini-3-Pro-preview
on per-sample hallucination and missing-fact counts.
Reference-conditioned judging improves agreement for both metrics.
}
\label{tab:app_judge_consistency_overall}
\begin{tabular}{lccccc}
\toprule
Evaluation mode & Pairs & $\rho_{\mathrm{hall}}$ & $p_{\mathrm{hall}}$ & $\rho_{\mathrm{miss}}$ & $p_{\mathrm{miss}}$ \\
\midrule
With reference & 480 & \textbf{0.537} & $2.1{\times}10^{-34}$ & \textbf{0.334} & $5.7{\times}10^{-13}$ \\
No reference & 480 & 0.377 & $2.5{\times}10^{-17}$ & 0.284 & $3.5{\times}10^{-10}$ \\
\bottomrule
\end{tabular}
\end{table}

\paragraph{Per-model consistency.}
Table~\ref{tab:app_judge_consistency_per_model} reports per-model
correlations. The SFT baseline obtains the highest judge agreement,
especially for hallucination:
$\rho_{\mathrm{hall}}{=}0.651$ with reference and $0.419$ without
reference. In contrast, the RL-trained models have lower agreement.
This pattern suggests that judge agreement is partly determined by
evaluation difficulty. The SFT baseline contains more obvious errors,
which both judges can identify consistently. After RL training,
captions become more accurate and the remaining errors are often
subtler, leading to lower judge correlation. Therefore, lower
per-sample agreement for RL checkpoints should not be interpreted as
worse model quality; rather, it indicates that the diagnostic task
becomes more difficult as captions improve.

\begin{table}[h]
\centering
\small
\caption{
Per-model Spearman correlations between GPT-5.2 and
Gemini-3-Pro-preview. RL-trained models show lower judge agreement,
suggesting that remaining errors become more subtle.
}
\label{tab:app_judge_consistency_per_model}
\resizebox{\textwidth}{!}{
\begin{tabular}{lcccccc}
\toprule
Model & Ref $\rho_{\mathrm{hall}}$ & No-ref $\rho_{\mathrm{hall}}$ & $\Delta_{\mathrm{hall}}$
& Ref $\rho_{\mathrm{miss}}$ & No-ref $\rho_{\mathrm{miss}}$ & $\Delta_{\mathrm{miss}}$ \\
\midrule
\textsc{ClaimDiff-RL} actor-only & 0.459 & 0.256 & $-0.203$ & 0.288 & 0.166 & $-0.122$ \\
\textsc{ClaimDiff-RL} relative & 0.452 & 0.321 & $-0.131$ & 0.270 & 0.192 & $-0.078$ \\
SFT baseline & \textbf{0.651} & \textbf{0.419} & $-0.232$ & \textbf{0.372} & \textbf{0.302} & $-0.070$ \\
\bottomrule
\end{tabular}
}
\end{table}

\paragraph{Takeaways.}
These analyses provide three supporting conclusions. First, Gemini's
claim-level judgments are reasonably reliable under human expert audit,
reaching $87\%$ per-claim accuracy on approximately $300$ audited
claims. Second, reference-conditioned judging improves GPT--Gemini
agreement compared with no-reference judging, validating the use of
references as comparison anchors rather than ground truth. Third,
judge agreement is higher for the SFT baseline than for RL-trained
models, suggesting that RL makes captions harder to evaluate because
the remaining errors are more subtle. This motivates reporting
aggregate trends and multiple diagnostic metrics rather than relying
on single-sample judgments.

\section{Ambiguity Penalty Ablation}
\label{app:ambiguity_ablation}

We ablate the ambiguity penalty in \textsc{ClaimDiff-RL} relative using the default severity weights \(w=(1,1.25,1.6)\). The ambiguity rate is computed as the number of matched ambiguity or hedging expressions from our parsing list, normalized by the number of generated words:
\[
\mathrm{Amb.}
=
\frac{\#\text{ambiguity phrase matches}}{\#\text{generated words}}
\times 100\%.
\]
As shown in Table~\ref{tab:app_ambiguity_ablation}, removing the ambiguity penalty increases the ambiguity rate from \(0.89\%\) to \(1.75\%\). This confirms that the penalty acts as a targeted safeguard against hedging rather than as a separate reward objective.

\begin{table}[h]
\centering
\small
\caption{
Ablation of ambiguity penalty on \textsc{ClaimDiff-RL} relative with default severity weights \(w=(1,1.25,1.6)\). MEDC is measured on the \(3\)K validation set; \(\overline{\mathrm{Hall}}\), \(\overline{\mathrm{Miss}}\), and ambiguity rate are measured on the human-labeled diagnostic benchmark.
}
\label{tab:app_ambiguity_ablation}
\begin{tabular}{lcccc}
\toprule
Reward variant 
& MEDC \(\downarrow\) 
& \(\overline{\mathrm{Hall}}\) \(\downarrow\) 
& \(\overline{\mathrm{Miss}}\) \(\downarrow\) 
& Amb. \(\downarrow\) \\
\midrule
\textsc{ClaimDiff-RL} relative 
& 0.68 & 1.52 & 0.80 & \textbf{0.89\%} \\
\quad w/o ambiguity penalty 
& 0.52 & 1.60 & 0.76 & 1.75\% \\
\bottomrule
\end{tabular}
\end{table}




\section{Code and Data Availability}
\label{app:code_data}

For anonymous review, we provide an anonymized supplementary repository at:
\url{https://anonymous.4open.science/r/ClaimDiff-RL-7486/}.
The repository includes the evaluation scripts, reward parsing code, prompt templates, diagnostic benchmark annotations, model outputs used in the reported tables and figures, and instructions for running the judge-based evaluation pipeline. After acceptance, we plan to release the code, data, and trained checkpoints publicly.


\end{document}